\documentclass[sigconf]{acmart}

\usepackage{graphicx}
\usepackage{subcaption}
\usepackage{multirow}
\usepackage{multicol}
\usepackage{colortbl}
\definecolor{mygray}{gray}{.9}
\usepackage{color}
\usepackage{balance} 

\AtBeginDocument{%
  }

\copyrightyear{2023}
\acmYear{2023} 
\setcopyright{acmlicensed}  
\acmConference[MM '23] {Proceedings of the 31st ACM International Conference on Multimedia}{October 29--November 3, 2023}{Ottawa, ON, Canada.}
\acmBooktitle{Proceedings of the 31st ACM International Conference on Multimedia (MM '23), October 29--November 3, 2023, Ottawa, ON, Canada}
\acmPrice{15.00}
\acmISBN{979-8-4007-0108-5/23/10}
\acmDOI{10.1145/3581783.3612477}

\acmSubmissionID{3360}

\begin{document}

\title{Learning Intra and Inter-Camera Invariance for Isolated Camera Supervised Person Re-identification}

\author{Menglin Wang}
\email{lynnwang6875@gmail.com}
\affiliation{%
  \institution{Xidian University}
  \city{}
  \state{}
  \country{}
}

\author{Xiaojin Gong}
\authornote{Corresponding author}
\email{gongxj@zju.edu.cn}
\affiliation{%
  \institution{Zhejiang University}
  \city{}
  \state{}
  \country{}}

\begin{abstract}
Supervised person re-identification assumes that a person has images captured under multiple cameras. However when cameras are placed in distance, a person rarely appears in more than one camera. This paper thus studies person re-ID under such isolated camera supervised (ISCS) setting. Instead of trying to generate fake cross-camera features like previous methods, we explore a novel perspective by making efficient use of the variation in training data. Under ISCS setting, a person only has limited images from a single camera, so the camera bias becomes a critical issue confounding ID discrimination. Cross-camera images are prone to being recognized as different IDs simply by camera style. To eliminate the confounding effect of camera bias, we propose to learn both intra- and inter-camera invariance under a unified framework. First, we construct style-consistent environments via clustering, and perform prototypical contrastive learning within each environment. Meanwhile, strongly augmented images are contrasted with original prototypes to enforce intra-camera augmentation invariance. For inter-camera invariance, we further design a much improved variant of multi-camera negative loss that optimizes the distance of multi-level negatives. The resulting model learns to be invariant to both subtle and severe style variation within and cross-camera. On multiple benchmarks, we conduct extensive experiments and validate the effectiveness and superiority of the proposed method. Code will be available at \textit{https://github.com/Terminator8758/IICI}.
\end{abstract}



\maketitle

\section{Introduction}

Person re-identification (re-ID) is the task of matching the same person in non-overlapping camera views. With its wide application in many areas of video surveillance, extensive studies have been performed which greatly pushed the advance of person re-ID. Supervised person re-ID has reached quite high accuracy~\cite{hermans2017defense, Zhai2019loss, chen2018video, sun2018beyond, zhou2019osnet, zheng2019dgnet, He2021vit}, mainly due to the development of deep learning networks~\cite{he2016deep, attention17, He2021vit} and availability of large-scale datasets~\cite{krizhevsky2012imagenet, 7410490, Wei2018PTGAN}. To alleviate the burden of data annotation, many researchers also focus on unsupervised re-ID. Effective unsupervised methods~\cite{ge2020mutual, ge2020self, Wang2021CAP, Chen2021ICE} are designed based on clustering and contrastive learning~\cite{xiao2017memory, wu2018memory, Chen2020SimCLR} paradigm, generating models that are even competitive against their supervised counterparts. However, both supervised and unsupervised re-ID require large-scale training set containing multi-view images per-person to be available. Collecting cross-camera images of the same person can be very tedious and difficult, especially when camera views are far apart. Therefore, it would be desirable if a re-ID model can learn from limited training data given only single-camera images of each person. This specific task is referred to as ``Isolated Camera Supervised (ISCS) re-ID", and has been studied in a few works~\cite{tian2020single, ge2021ccfp, Wu2022ccsfg} previously.

Zhang \textit{et. al}~\cite{tian2020single} are the first to propose such re-ID setting. To alleviate the influence of camera style bias on cross-camera recognition, they design a multi-camera negative loss that optimizes the distance ranking of intra- and cross-camera negative pairs. The resulting model learns to discriminate cross-camera images with less camera bias involved. Later works CCFP~\cite{ge2021ccfp} and CCSFG~\cite{Wu2022ccsfg} both resort to cross-camera feature generation to make up for the absence of cross-camera positive images. CCFP adopts camera-specific BNs~\cite{Sergey2015}, while CCSFG~\cite{Wu2022ccsfg} proposes camera-conditioned variational auto-encoder to generate cross-camera fake positive features. Nevertheless, the generated feature may be semantically-noisy or lack diversity in pedestrian view or motion, hindering model performance from being further improved.

\begin{figure}[ht]
\centering
\begin{subfigure}{0.39\textwidth}
\centering
\includegraphics[width=1.0\textwidth]{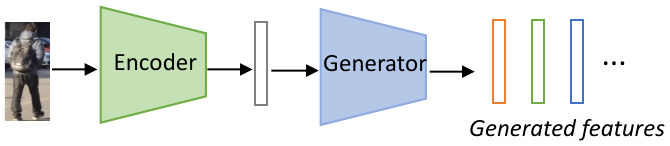} 
\caption{Previous methods: cross-camera feature generation}
\end{subfigure}
\\
\begin{subfigure}{0.43\textwidth}
\centering
\includegraphics[width=1.0\textwidth]{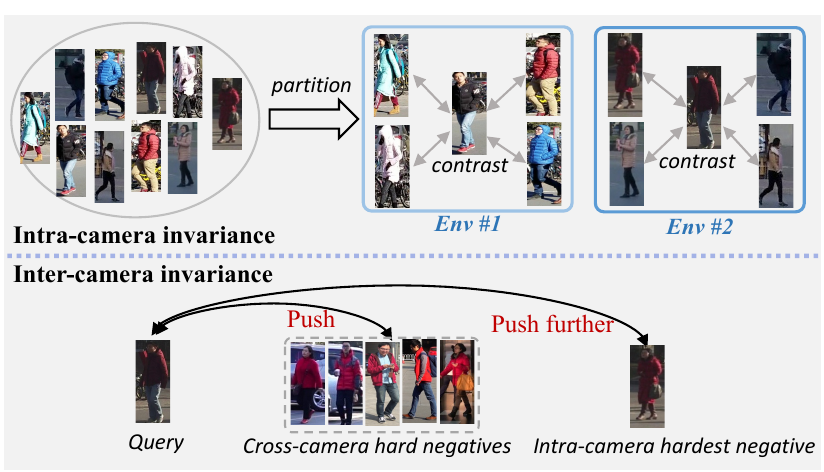} 
\caption{Our method: learning invariance within data}
\end{subfigure}
\caption{A basic idea of our intra-camera and inter-camera invariance learning.}
\label{fig_intro}
\end{figure}

Instead of explicitly generating synthetic features, in this work we propose a novel perspective to address the ISCS re-ID problem. Considering that camera style bias is the key factor confounding cross-camera ID discrimination. We exploit the variation in the given training data, and aim to learn a robust model by enforcing invariance both within and cross-camera. A preliminary idea of our invariance learning is shown in Figure \ref{fig_intro}.

Specifically, each camera view provides a natural sub-domain with roughly consistent style (eg. illumination, contrast). With style variance eliminated, we can safely train a model that learns to discriminate different intra-camera IDs without taking camera style as a shortcut, thus achieving a basic level of intra-camera invariance. To this end, we adopt prototypical contrastive learning where images are contrasted with intra-camera class prototypes following InfoNCE~\cite{InfoNCE10, wu2018memory}. To make the invariance learning more effective, we further take both weakly and strongly (\textit{eg.} color-jittered) augmented images with asymmetric prototype update, to enforce perturbation robustness and augmentation invariance.

Sometimes, there may exist subtle style variation within camera, so intra-camera learning could still be confounded by style bias. In this scenario, we resort to intra-camera clustering to generate style-consistent environments. This is achieved by exploiting low-layer backbone feature whose statistics encode style-related information~\cite{huang017iccv, zhou2021mixstyle}. Clustering on such feature produces sub-cameras each of which contains style-consistent IDs. By performing contrastive learning within sub-camera environments, the model is able to bypass the shortcut bias~\cite{Geirhos2020} during intra-camera learning. 

In order to further improve cross-camera discrimination, the model still needs to compare and recognize cross-camera IDs. MCNL loss~\cite{tian2020single} provides a delicate way of modulating the relation of intra and inter-camera negatives, so that the model can discriminate cross-camera IDs without being too much influenced by camera bias. However, it only considers the hardest negative cross-camera, which kind of weakens its effect. In this work, we take a step further and extend MCNL loss from two aspects. First, multiple cross-camera hardest negatives are taken for optimization, and second, multi-level loss is considered to compare with not only in-batch instances but also global class prototypes. These two modifications ensure the model to be much more resistant to camera style interference. With both intra- and inter-camera invariance learning, the model learns to discriminate IDs regardless of their camera style, thus gaining robustness to camera bias. 

The main contributions of our method are summarized as follow:
\begin{itemize}
\item {We propose a unified framework that learns intra- and inter-camera invariance to address the camera bias issue for ISCS re-ID task.}
\item {To learn intra-camera invariance, we improve vanilla prototypical contrastive learning with augmentation invariant contrast as well as sub-camera environment generation. On this basis, a multi-level multi-negative loss is proposed which strengthens the effect of inter-camera style invariance.}
\item {We evaluate the proposed method on two benchmark datasets with varying backbones, and extensive experiments demonstrate clear improvements over previous state-of-the-arts.}
\end{itemize}

\section{Related work}

\subsection{Intra-Camera Supervised Re-ID}
Intra-camera supervised (ICS) re-ID setting is formally proposed in \cite{zhu2019intra}, as an attempt to reduce the data annotation burden in person re-ID. ICS re-ID only requires identity labels to be independently labeled within each camera view, while inter-camera ID relation is unknown. Considering its characteristics, existing works usually integrate two components: intra-camera supervised learning to make full use of ground truth annotation, and cross-camera association to discover cross-camera ID relations for model learning. Specifically, MTML~\cite{zhu2019intra} adopts multi-branch parametric classifiers for supervised intra-camera classification, then predicts cross-camera association by cycle consistency. In their extension work~\cite{zhu2020intra}, a curriculum cycle association is designed by thresholding the identity joint matching probability. PCSL~\cite{qi2019progressive} and ACAN~\cite{qi2019intra} both adopt the triplet loss for intra-camera learning, and utilize multi-camera adversarial learning~\cite{qi2019intra} or nearest-neighbor soft-labeling~\cite{qi2019progressive} for inter-camera learning. Precise-ICS~\cite{Wang2021ICS} adopts per-camera non-parametric classifiers along with a quintuplet loss for intra-camera learning, and formulates cross-camera ID association as a problem of finding connected components in a graph.

\subsection{Isolated-Camera Supervised Re-ID}
Isolated-camera supervised re-ID~\cite{tian2020single} is a similar setting to ICS re-ID, but each person in the training set only appears in one camera. 
As an effort to low-cost data collection, this setting also raises new challenges for model learning. Since there is no cross-camera positive images, model learning is inevitably impacted by the camera style bias. The model tends to learn easy features like camera style as a shortcut~\cite{Geirhos2020} to recognize IDs. To mitigate this,  Zhang \textit{et. al}~\cite{tian2020single} design a multi-camera negative loss that ensures the most similar negative image is found in other cameras. CCFP~\cite{ge2021ccfp} and CCSFG~\cite{Wu2022ccsfg} try to generate synthetic features to complement the lack of cross-camera positive images. In specific, CCFP~\cite{ge2021ccfp} utilizes camera-specific batch normalization, while CCSFG~\cite{Wu2022ccsfg} adopts camera-conditioned variational encoder to synthesize cross-camera positive features. After obtaining the synthetic features, they compute a consistency loss to minimize the distance of those cross-camera positive pairs. Instead of generating synthetic features, we take a different perspective and regard the task as an invariance learning problem. By exploiting the variation within and cross-camera, our method can better deal with the camera style shift and remove the shortcut bias during the invariance learning.

\begin{figure*}[ht]
\centering
\includegraphics[width=0.85\textwidth]{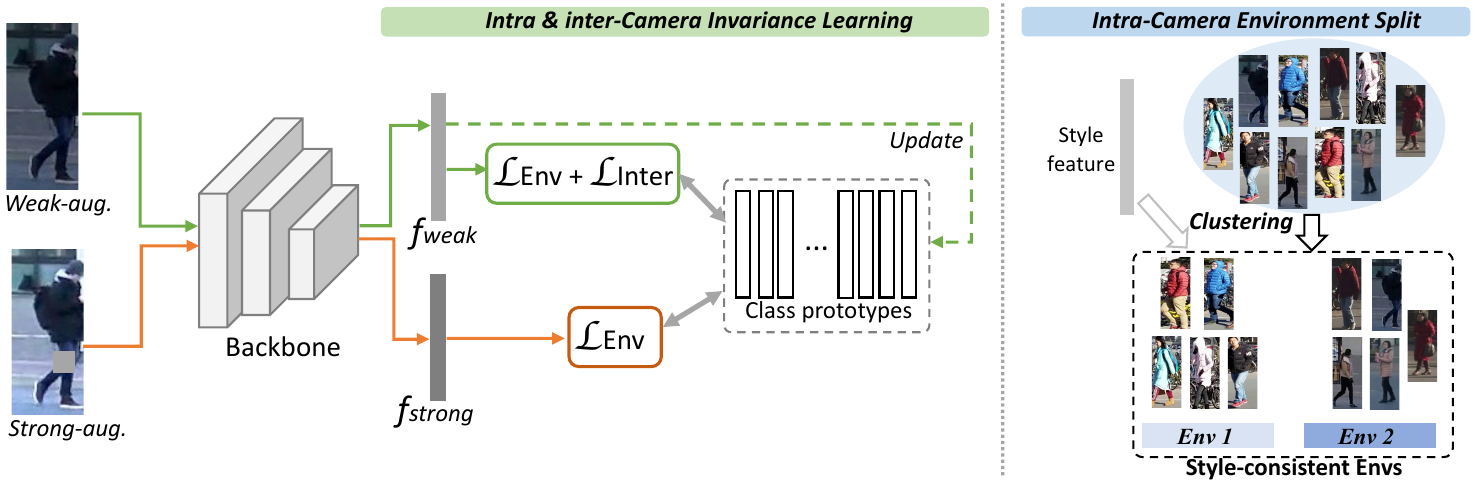} 
\caption{An overview of our proposed method. During training, both weakly and strongly-augmented images are forwarded by the backbone to obtain their representations. For weak-augmented images, $\mathcal{L}_{Env}$ and $\mathcal{L}_{Inter}$ are computed. During backward, class prototypes are updated by features of weak-augmented images. For strong-augmented images, only $\mathcal{L}_{Env}$ is computed to enforce perturbation invariance. To generate style-consistent environments, intra-camera ID clustering is performed with low-layer feature which preserves style information.}
\label{fig_framework}
\end{figure*}

\subsection{Invariance Learning}
In person re-ID task, ECN~\cite{zhong2019invariance, 9018132} learns three types of invariance for domain adaptive re-ID. In their method, exemplar invariance enforces each exemplar away from others, camera invariance overcomes camera style variation by augmenting style transfer images, and neighborhood invariance draws pseudo positive images close to each other. Chen \textit{et. al}~\cite{9577603, chen2023} enhance view-invariance by maximizing the agreement between original, synthesized, and memory representations. 
In domain generalization, some works~\cite{muandet2013, 8578664, li2018invariant} also learn domain-invariant representation with multiple diverse source domains, based on the idea of aligning features of different domains. EqInv~\cite{wang2022equivariance} performs both equivariant feature learning and invariance risk minimization for learning from insufficient data. Our method also embraces the idea of invariance learning for ISCS re-ID task, where intra-camera variance bypasses the influence of camera bias, and inter-camera variance directly confronts it.

\section{Methodology}

\subsection{Overview}
The proposed method is termed \textbf{IICI} (``learning Intra- and Inter-Camera Invariance"), for isolated camera supervised re-ID task. Given a multi-camera training set with each person appearing only in one camera, we aim to learn a re-ID model that can well discriminate different IDs and perform robustly in test-time cross-camera retrieval. We approach this by exploiting the variation in the given training set, and enhance the model's discrimination ability through intra- and inter-camera invariance learning. An overview of the proposed method is illustrated in Figure \ref{fig_framework}.

\subsection{Intra-Camera Invariance Learning}

\subsubsection{\textbf{An intra-camera learning baseline.}}
A critical issue of ISCS re-ID is the camera shortcut bias caused by the lack of cross-camera positive pairs. If directly classifying IDs in all cameras, the model easily learns the camera style difference as a shortcut to recognize IDs from different camera views. This originates from the characteristics of feature extraction in deep network. Deep models tend to stop learning once simple feature like image style suffices for loss optimization~\cite{wang2022equivariance}.  Instead of global ID classification, we find that intra-camera learning is a good alternative to avoid the shortcut learning problem. Different camera views create natural sub-domains, where images usually exhibit similar background and image style. So when classifying images under the same camera, the model can no longer use the camera style as a shortcut to optimize the loss. It is thus forced to learn higher-level semantic features to recognize different intra-camera IDs. 

We first introduce a baseline for intra-camera learning. Given a training set $\mathcal{D} = \{(x_i, y_i, c_i) \}_{i=1}^{N}$ where $x_i$ is the $i$-th image, $y_i$ and $c_i$ are respectively the identity label and camera label of $x_i$, $N$ is the number of images. Suppose there are $Y$ identities and $C$ camera views in training set. For ease of notation, we also define $\mathcal{D}_c$ as the image subset of camera c, and $\mathcal{Z}_c$ as the set of unique identity labels from camera $c$. Following the successful application~\cite{Wang2021ICS, ge2020self, Wang2021CAP} of non-parametric contrastive learning in re-ID, we adopt the prototypical contrastive learning paradigm. Specifically, we denote the backbone network as $f_\theta$. When image $x_i$ is input, the network extracts a $d$-dimensional $L$2-normalized feature as $f_\theta(x_i)$. Then we construct a feature memory bank $\mathcal{M} \in R^{Y \times d}$ whose each entry represents a class prototype. During backward, the memory feature is updated by online batch features in a moving average manner:
\begin{equation}
\mathcal{M}[y_i] \leftarrow \mu \mathcal{M}[y_i] + (1 - \mu) f_\theta(x_i),
\label{eq:mu}
\end{equation}

where $\mathcal{M}[y_i]$ is the $y_i$-th entry of the memory bank, also representing the $y_i$-th class prototype. $\mu \in [0,1]$ is an updating rate. After each update, $\mathcal{M}[y_i]$ is $L$2-normalized to have unit norm.

The baseline intra-camera loss is then defined as:
\begin{equation}
\mathcal{L}_{Intra1} = - \sum_{c=1}^{C} \frac{1}{|\mathcal{D}_c|} \sum_{x_i \in \mathcal{D}_c} \log \frac{exp(\mathcal{M}[y_i]^T f_\theta(x_i)/\tau)}{\sum_{j \in \mathcal{Z}_c} exp(\mathcal{M}[j]^T f_\theta(x_i)/\tau)},
\label{eq_intra1}
\end{equation}
where $\tau$ is the temperature, $\mathcal{M}[j]^T f_\theta(x_i) / \tau$ represents the scaled similarity between image feature $f_\theta(x_i)$ and class prototype $\mathcal{M}[j]$.

\subsubsection{\textbf{Augmentation-invariant contrast.}}
The above intra-camera loss $\mathcal{L}_{Intra1}$ is only computed on ``weakly-augmented" input images. It preserves the original consistent style under each camera, in order to keep the model focused on learning ID-relevant features for discrimination. Nevertheless, it also limits the model to perceive only one single style during contrast. To let the model adapt to more diverse styles and be more robust to perturbation, we propose to enforce augmentation-invariant contrast. During mini-batch training, we apply both two different augmentations (weak, strong) to each in-batch image. The ``strongly-augmented" (\textit{with color jittering and stronger random erasing}) batch images are denoted as $\{x_1^{aug}, x_2^{aug}, ...\}$. Then we compute another intra-camera contrastive loss \textit{w.r.t.} the strongly-augmented images:
\begin{equation}
\small
\mathcal{L}_{Intra2} = - \sum_{c=1}^{C} \frac{1}{|\mathcal{D}_c|} \sum_{x_i \in \mathcal{D}_c} \log \frac{exp(\mathcal{M}_{sg}[y_i]^T f_\theta(x_i^{aug})/\tau)}{\sum_{j \in \mathcal{Z}_c} exp(\mathcal{M}_{sg}[j]^T f_\theta(x_i^{aug})/\tau)},
\label{eq_intra2}
\end{equation}

Here, we use $\mathcal{M}_{sg}[j]$ to represent gradient-detached $\mathcal{M}[j]$. This means that \textit{only weakly-augmented features are used to update class prototypes in an asymmetrical manner.} Therefore, class prototypes retain the original style, and the contrast in Eq. (\ref{eq_intra2}) is between style-jittered image features and original-style prototypes. The model is enforced to maximize the invariance between style-jittered feature and its positive class prototype when optimizing Eq. (\ref{eq_intra2}), thus achieving a higher level of perturbation robustness. 

\textit{\textbf{Analysis:}} An alternative to exploit the style-shifted features is to directly compute a consistency loss~\cite{ge2021ccfp, Wu2022ccsfg} that minimizes the distance between original and style-shifted features. We empirically find this to be not as effective as Eq. (\ref{eq_intra2}). Another possibility is to directly apply the strong augmentations to input batch, so the class prototypes will also be updated with strongly-augmented features. As shown in our experiments, this gives inferior performance than $\mathcal{L}_{Intra1}+\mathcal{L}_{Intra2}$, indicating that keeping class prototypes with original style is important.

\subsubsection{\textbf{Intra-camera environment split.}}
For some datasets, there exist subtle style variation within certain camera, hampering the model learning from being invariant to image style. For example, under the same camera view, person images could have varying backgrounds or illumination condition due to different time slots. Compared to cross-camera variation, intra-camera variation is less severe but non-negligible. Attending to intra-camera variance helps the model to focus on learning foreground-related semantics and enables stronger invariance guarantees. To this end, we propose to split those style-varying cameras into sub-cameras. Each sub-camera should form a style-consistent environment, so that intra-camera invariance learning can fully eliminate style interference.

\begin{figure}[h]
\centering
\includegraphics[width=0.33\textwidth]{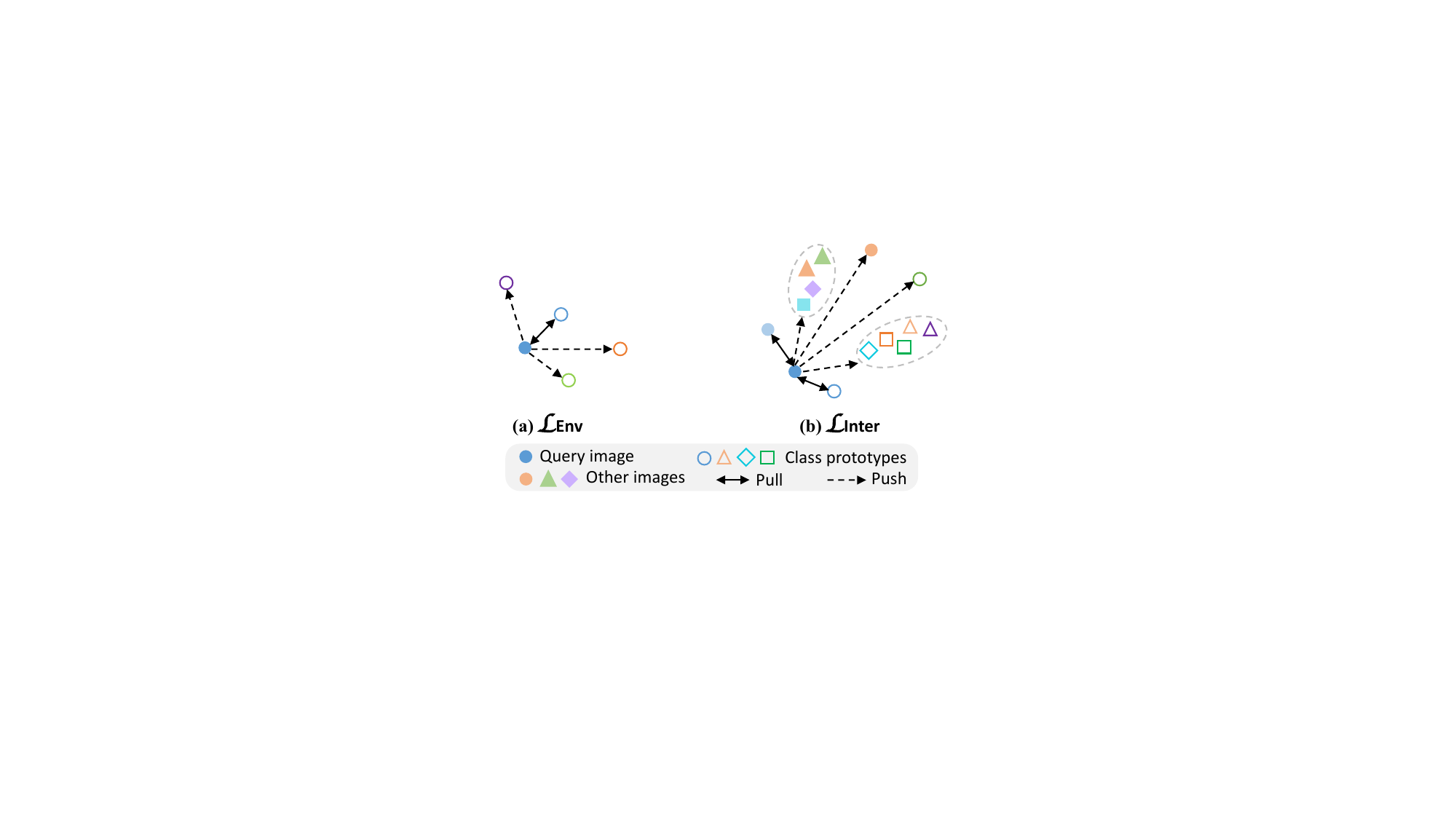} 
\caption{Illustration of $\mathcal{L}_{Env}$ and $\mathcal{L}_{Inter}$. Different shape (eg. circle, triangle, square) represents different environment, different color represents different ID. Solid shapes are images.}
\label{fig_loss}
\end{figure}

We exploit intra-camera ID clustering to generate sub-camera environments. According to previous works~\cite{huang017iccv, vicent2017iclr, zhou2021mixstyle}, low-layer network feature statistics often encode style-related information. Therefore, we use the low-layer pooled feature to perform ID clustering. Specifically, K-Means is adopted to split IDs from one camera into two or more sub-camera environments, depending on the ID number under the specific camera. To update the environments along with model training, we perform K-means epoch-wise. Suppose after ID clustering, $SC$ sub-cameras are generated. For each sub-camera $u$, we denote its image set as $\mathcal{D}_u^{'}$, and its unique identity label set as $\mathcal{Z}_u^{'}$. Then we revise the intra-camera invariance learning as intra-environment invariance learning, as follow:
\begin{equation}
\mathcal{L}_{Intra1}^{'} = - \sum_{u=1}^{SC} \frac{1}{|\mathcal{D}_u^{'}|} \sum_{x_i \in \mathcal{D}_u^{'}} \log \frac{exp(\mathcal{M}[y_i]^T f_\theta(x_i)/\tau)}{\sum_{j \in \mathcal{Z}_u^{'}} exp(\mathcal{M}[j]^T f_\theta(x_i)/\tau)},
\label{eq_intra1_new}
\end{equation}

Likewise, $\mathcal{L}_{Intra2}^{'}$ is also defined within sub-camera. The total loss for intra-environment learning is then defined as:
\begin{equation}
\mathcal{L}_{Env} = \mathcal{L}_{Intra1}^{'} + \mathcal{L}_{Intra2}^{'},
\label{eq_env}
\end{equation}

\subsection{Inter-Camera Invariance Learning}
Intra-camera invariance learning avoids the interference of camera bias by fixing the image style variable. Nevertheless, cross-camera  learning is still necessary for the model to recognize IDs in different cameras more accurately.

\subsubsection{\textbf{The MCNL loss.}} Proposed in \cite{tian2020single}, the MCNL loss is proven effective for cross-camera learning in ISCS re-ID. It constrains cross-camera hard negative pair to be closer than intra-camera hard negative pair, forcing the model to learn style-invariant features in order to optimize the loss. Specifically, the loss is written as:
\begin{equation}
\begin{aligned}
& \mathcal{L}_{MCNL} = \frac{1}{N} \sum_{i=1}^N [m_1 + d(f_i, f_{intra}^+) - d(f_i, f_{cross}^-)]_+  \\
&\ \ \ \ \ \ \ \ \ \ \ \ \ \ \ + [m_2 + d(f_i, f_{cross}^-) - d(f_i, f_{intra}^-)]_+,
\end{aligned}
\label{eq_mcnl}
\end{equation}
where $m_1$ and $m_2$ are two margins empirically set as 0.1, and $[z]_+=max(z, 0)$. $d(\cdot)$ is the euclidean distance, $f_i$ is a short notation of $f_\theta(x_i)$, $f_{intra}^+$ is the intra-camera hardest positive in batch, $f_{intra}^-$ is the intra-camera hardest negative in batch, and $f_{cross}^-$ is the cross-camera hardest negative in batch.

\subsubsection{\textbf{Our improved inter-camera loss.}}
We note that the original MCNL loss only optimizes the hardest cross-camera negative instance. It does not fully exploit the potential within other cross-camera hard negatives that might contain complementary information to the hardest negative. Hence, we make the first attempt to improve the original MCNL loss in two perspectives. An illustration of the improved loss is shown in Figure \ref{fig_loss} (b). 

First, we not only consider the hardest cross-camera negative, but also \textbf{make use of the top-K cross-camera hard negatives}:
\begin{equation}
\begin{aligned}
& \mathcal{L}_{Inter1} = \frac{1}{N} \sum_{i=1}^N \sum_{k=1}^{K_1} [m_1 + d(f_i, f_{intra}^+) - d(f_i, f_{cross,\textbf{k}}^-)]_+ \\
&\ \ \ \ \ \ \ \ \ \ \ \ \ \ \ + [m_2 + d(f_i, f_{cross,\textbf{k}}^-) - d(f_i, f_{intra}^-)]_+, 
\end{aligned}
\label{eq_mcnl2}
\end{equation}
where $f_{cross,\textbf{k}}$ is the $k$-th cross-camera hardest negative. $K_1$ is the number of selected top-K cross-camera hard negative instances.

As shown in Eq. (\ref{eq_mcnl2}), $\mathcal{L}_{Inter1}$ is computed for top $K_1$ cross-camera hard negatives in the batch. The motivation is that due to the much larger number of cross-camera identities, it's \textit{relatively easy} to guarantee the hardest cross-camera image to have higher similarity than the hardest intra-camera image. TopK cross-camera hard negatives represent more diverse distribution of cross-camera styles, so when taking TopK cross-camera hard negatives into comparison, the effect of MCNL loss can be much strengthened. The original MCNL loss $\mathcal{L}_{MCNL}$ can be seen as a special case of $ \mathcal{L}_{Inter1}$ when $K_1$ is 1.

Next, taking advantage of the global class prototypes in our framework, we also enforce a \textbf{class prototype based loss}:
\begin{equation}
\begin{aligned}
& \mathcal{L}_{Inter2} = \frac{1}{N} \sum_{i=1}^N \sum_{k=1}^{K_2} [m_1 + d(f_i, \mathcal{M_{sg}}[_{intra}^+]) - d(f_i, \mathcal{M}_{sg}[_{cross,\textbf{k}}^-])]_+   \\
&\ \ \ \ \ \ \ \ \ \ \ \ \ \ \ + [m_2 + d(f_i, \mathcal{M}_{sg}[_{cross,\textbf{k}}^-]) - d(f_i, \mathcal{M}_{sg}[_{intra}^-])]_+, 
\end{aligned}
\label{eq_mcnl3}
\end{equation}
where $K_2$ is the number of selected top-K cross-camera hard negative prototypes.

$\mathcal{L}_{Inter2}$ compares each in-batch image with all the class prototypes globally. It breaks the limit of batch size, and mines more diverse candidates containing possibly more similar negatives. Like $\mathcal{L}_{Inter1}$, it considers $K_2$ hardest cross-camera negative prototypes for optimization. 

The inter-camera loss is the combination of $\mathcal{L}_{Inter1}$ and $\mathcal{L}_{Inter2}$:
\begin{equation}
\mathcal{L}_{Inter} = \mathcal{L}_{Inter1} + \mathcal{L}_{Inter2}, 
\label{eq_inter}
\end{equation}

It eliminates the influence of cross-camera style variation, facilitating the model in being more robust and invariant to style change across-camera. Note that only weakly-augmented images in the batch are adopted for inter-camera loss.

With intra-camera and inter-camera invariance learning, the final loss is formulated as:
\begin{equation}
\mathcal{L}_{Overall} = \mathcal{L}_{Env} + \mathcal{L}_{Inter}.
\label{eq_overall}
\end{equation}

\section{Experiments}

\subsection{Datasets and Evaluation Metrics}
We evaluate the proposed method on subsets of two benchmark re-ID datasets, Market-1501~\cite{7410490} and MSMT17~\cite{Wei2018PTGAN}.  
Following \cite{ge2021ccfp, Wu2022ccsfg}, the camera-isolated training set is generated by randomly selecting images of one camera view for each identity in the original training set, to make sure each identity only appear in a single camera. The testing set, including gallery and query set, are kept unchanged. The resulting camera-isolated datasets are referred to as \textbf{Market-SCT} and \textbf{MSMT-SCT}, respectively. The detailed statistics of the generated datasets, as well as original datasets, are listed in Table \ref{dataset_statistic_table}.

For evaluation metrics, we adopt the commonly used mean Average Precision (mAP) and Cumulative Matching Characteristic (CMC). Specifically, the Rank-1, Rank-5 and Rank-10 of the CMC metric are reported along with mAP. For fair comparison, we do not use any post-processing techniques such as re-ranking~\cite{Zhong2017reranking} during evaluation.

\begin{table}[h]
\caption{Statistics of each dataset and their camera-isolated version.}
\centering
\scalebox{0.85}{
\begin{tabular}{l|ccccc}
\hline  
\multirow{2}{*}{Dataset} & \multirow{2}{*}{\shortstack{\texttt{\#}Train\\ IDs}} & \multirow{2}{*}{\shortstack{\texttt{\#}Training\\ Images}} 
& \multirow{2}{*}{\shortstack{\texttt{\#}Test\\ IDs}} & \multirow{2}{*}{\shortstack{\texttt{\#}Test\\ Images}} & \multirow{2}{*}{\shortstack{\texttt{\#}Has cross-cam\\ person (train)}}\\
\\  \hline 
Market-1501         & 751   & 12,936  & 750     & 15,913    & True \\
Market-SCT         & 751   & 3,561  & 750     & 15,913       & False\\
MSMT17              & 1,041 & 32,621  & 3,060  & 93,820   & True \\ 
MSMT-SCT              & 1,041 & 6,694  & 3,060  & 93,820  & False \\ 
\hline
\end{tabular}
}
\label{dataset_statistic_table}
\end{table}

\subsection{Implementation Details}
We conduct experiments on three backbones: 1) \textbf{ResNet50-Nonlocal}. This is the backbone utilized by both CCFP~\cite{ge2021ccfp} and CCSFG~\cite{Wu2022ccsfg}. Following their implementation, we also optionally add a DeTR-adapted~\cite{carion2020detr} local branch after the final ResNet block, to extract an auxiliary feature. All of our ablation studies are performed using this backbone.
2) \textbf{ResNet50-IBN}~\cite{he2016deep, Pan2018ibn}. This is a backbone commonly utilized for person re-ID, so we also present our full model result using this backbone. 3) \textbf{ViT-S}~\cite{He2021vit, dosovitskiy2021image} pre-trained on LUPerson~\cite{Zhong2017reranking}. We use backbone mainly to investigate how compatible our method is with pure transformer architecture.

When using ResNet50-based backbone, parameters are pre-trained on ImageNet, then we discard the classification layer and add a Batch Normalization layer right after the global pooling layer, following \cite{luo2019trick, ge2021ccfp}. Generalized mean pooling~\cite{ge2021ccfp, Wu2022ccsfg, raden2018finetuning} is utilized to perform global feature pooling. The 2048-dimensional output feature is then L2-normalized and used for loss computation as well as test-time evaluation. During training, batch size is 64, PK sampling strategy is adopted to sample 16 random persons and 4 images per-person in a mini-batch. Each sampled image is applied two data augmentations to obtain two transformed images: the first is \textit{weak augmentation}, including random flipping, cropping, and erasing. The second is \textit{strong augmentation}, including additional color jittering and stronger random erasing. The model is trained by Adam optimizer~\cite{Kingma2014} for 100 epochs, with the learning rate initially set to 0.00035 and divided by 10 every 20 epochs. As for the hyper-parameters, the memory update rate $\mu$ is $0.2$, the temperature factor $\tau$ is $0.05$, number of topK hardest negatives $K_1$ is set as 10. $K_2$ is set as 20 for Market-SCT and 50 for MSMT-SCT.

When using the ViT-S backbone, SGD optimizer is adopted with batch size set as 256 following \cite{He2021vit}. The total number of epochs is 50. Other hyper-parameters are similar as the ResNet50 backbone.

\subsection{Ablation Study}
In this section, we perform ablation study on each of our proposed components, to validate their effectiveness. The ablation results are summarized and presented in Table \ref{ablation_table}.

\subsubsection{\textbf{Effectiveness of our intra-camera baseline $\mathcal{L}_{Intra1}^{'}$.}}
First we analyze the baseline of our intra-camera invariance learning, represented by $\mathcal{L}_{Intra1}^{'}$ in Table \ref{ablation_table}.
Before that, we first define a global contrastive loss as $\mathcal{L}_{Base} = -\frac{1}{N} \sum_{i=1}^{N} \log \frac{exp(\mathcal{M}[y_i]^T f_\theta(x_i)/\tau)}{\sum_{j=1}^Y exp(\mathcal{M}[j]^T f_\theta(x_i)/\tau)}$.
The difference between $\mathcal{L}_{Base}$ and $\mathcal{L}_{Intra1}^{'}$ is that the former is contrasted with all the class prototypes, while the latter only contrasts intra-camera (or intra sub-camera) prototypes.

Compared to model $A1$ that adopts $\mathcal{L}_{Base}$, $A2$ utilizes intra-camera contrastive loss and achieves higher accuracy. On Market-SCT and MSMT-SCT, rank-1 is improved by $26.9\%$ and $32.8\%$ respectively. This proves that intra-camera loss serves as a better baseline for ISCS re-ID, since it avoids the confusion and shortcut brought by camera style variation. On the contrary, global supervised contrastive loss takes all IDs into classification, and cross-camera comparison is inevitably influenced by the misleading camera style difference.
Note that for MSMT-SCT, the intra-loss is computed within the sub-camera environment, in order to eliminate the style variation more thoroughly. An analysis on sub-camera environment split can be found in section $4.4.3$. 

\subsubsection{\textbf{Effectiveness of augmentation-invariant contrast $\mathcal{L}_{Intra2}^{'}$.}}
Built on the intra-camera baseline, the augmentation-invariant contrast also plays an important role in intra-camera invariance learning. In Table \ref{ablation_table}, $A3$ directly applies strong augmentation to batch input, and achieves higher accuracy than $A2$. Still, the combination of $\mathcal{L}_{Intra1}^{'}$ and $\mathcal{L}_{Intra2}^{'}$, as indicated by $A4$, is able to outperform both $A2$ and $A3$. For example on MSMT-SCT, rank-1 and mAP are boosted by $3.4\%$ and $1.9\%$. This proves the efficacy of contrasting both weakly and strongly-augmented inputs with original weakly-augmented prototypes, and also shows that enforcing augmentation-based contrast is beneficial to intra-camera invariance learning.

\subsubsection{\textbf{Effectiveness of the inter-camera loss.}}
For inter-camera learning, we propose a combination of two inter-camera losses: topK extended in-batch loss $\mathcal{L}_{Inter1}$ and prototype-based loss $\mathcal{L}_{Inter2}$. Now we analyze how the two losses contribute to model learning. In Table \ref{ablation_table}, $A5$ is trained with $\mathcal{L}_{Inter1}$ at $K_1$=1, which equals to the vanilla MCNL loss. Comparing $A5$ with $A6$, it is clear that considering multiple cross-camera hard negatives is more effective and leads to superior performance. Moreover, by comparing $A7$ and $A8$, we can see the global prototype-based $\mathcal{L}_{Inter2}$ further improves the performance by optimizing the cross-camera distance relation between images and class prototypes. Specifically, rank-1 is improved by $1.5\%$ on Market-SCT and $2.1\%$ on MSMT-SCT. It proves that $\mathcal{L}_{Inter2}$ is a useful complement to $\mathcal{L}_{Inter1}$, and the two losses both contribute to the effectiveness of inter-camera invariance learning.

\subsubsection{\textbf{Effectiveness of the full model.}}
Finally, we analyze the effectiveness of the proposed full model, indicated by $A9$ and $A10$ in Table \ref{ablation_table}. Combining all the proposed components, $A9$ gives better performance than $A1$-$A8$ for both Market-SCT and MSMT-SCT. When further adding a batch inter-camera loss on DeTR branch feature, the performance continues to improve on Market-SCT. On MSMT-SCT, the mAP is improved but rank-1 is decreased a bit, indicating that the DeTR branch is not necessarily useful in all cases.

\begin{table}[ht]
\centering
\caption{Comparison of the proposed method and its variants. $\mathcal{L}_{Intra1}^{'}$ and $\mathcal{L}_{Intra2}^{'}$ are computed under original camera for Market-SCT, and under sub-camera environment for MSMT-SCT. \textit{r.+d.} means computing $\mathcal{L}_{MCNL}^{'}$ on both ResNet feature and DeTR branch feature.}
\scalebox{0.85}{
\begin{tabular}{c|cccc|cc|cc}
\hline  
\multirow{2}{*}{} & \multicolumn{4}{c|}{Components} &\multicolumn{2}{c|}{Market-SCT}  & \multicolumn{2}{c}{MSMT-SCT}   \\
\cline{2-9}  & $\mathcal{L}_{Intra1}^{'}$  &  $\mathcal{L}_{Intra2}^{'}$ & $\mathcal{L}_{Inter1}$  & $\mathcal{L}_{Inter2}$ & R1  & mAP   & R1  & mAP  \\ 
\hline
$A1$        &  \multicolumn{4}{c|}{Baseline $\mathcal{L}_{Base}$}     & 54.5 & 37.6          & 18.3  & 8.6             \\
$A2$        & \checkmark &  &  &                                        & 81.4 & 63.5          & 51.1  & 23.4             \\
$A3$       &  & \checkmark &  &                                        & 82.6 & 64.7          & 53.3  & 24.4         \\
$A4$       &  \checkmark & \checkmark &  &                    & 82.7 & 65.4          & 54.5  & 25.3         \\
\hline
$A5$       &  &  & $K_1$=1 &                                        & 60.6 & 37.4          & 18.3  & 7.8             \\
$A6$       &  &  & \checkmark &                                        & 71.5 & 47.1          & 24.4  & 9.7             \\
$A7$        &  \checkmark &  & \checkmark &                    & 83.9 & 66.7          & 52.4  & 23.8         \\
$A8$        &  \checkmark &  & \checkmark &  \checkmark   & 85.4 & 67.5          & 54.5  & 24.6         \\
\hline 
$A9$       & \checkmark & \checkmark &\checkmark & \checkmark & 86.0 & 68.4          & 56.2  & \textbf{26.0}     \\
$A10$       & \checkmark   & \checkmark   & \textit{r.+d.}   & \checkmark   & \textbf{87.2} & \textbf{69.4 }          & \textbf{56.9}  & 24.6 \\
\hline
\end{tabular}
}
\label{ablation_table}
\end{table}

\begin{figure*}[ht]
\centering
\includegraphics[width=0.83\textwidth]{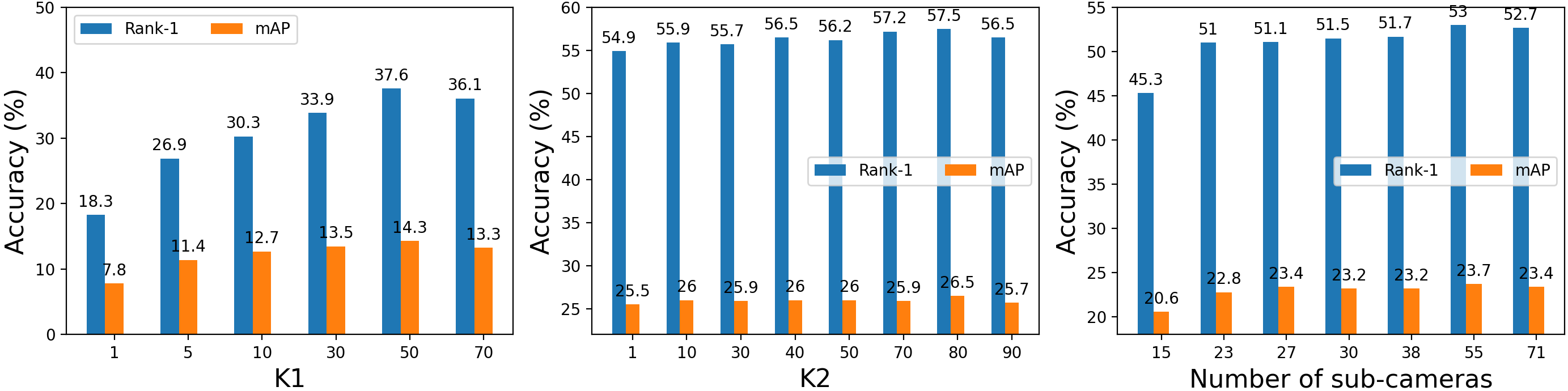} 
\caption{Analysis of hyper-parameters $K1$, $K2$ and number of sub-cameras on MSMT-SCT dataset.}
\label{fig_k1_k2_subcam}
\end{figure*}

\subsection{Parameter Analysis}

\subsubsection{\textbf{Number of cross-camera hard negative instances $K_1$}}
In Eq. (\ref{eq_mcnl2}), $K_1$ cross-camera hard negatives are selected for in-batch inter-camera loss. To investigate how the value of $K_1$ affects the performance, we conduct experiments with different values of $K_1$, and present the results in Figure \ref{fig_k1_k2_subcam} (a). By observing the figure, we can see the effectiveness of considering top-K cross-camera hard negatives. K1=1 represents the original MCNL loss. Simply increasing K1 to 5 already brings clear improvement in accuracy. Further increasing K1 allows more hard negatives of diverse styles to be optimized, which leads to much higher accuracy than the original MCNL loss. The optimal value of K1 is influenced by the batch size, and typically a larger batch size offers more candidates to be chosen from. When using a batch size of 128 as in Figure \ref{fig_k1_k2_subcam} (a), the optimal K1 saturates around 50.

\subsubsection{\textbf{Number of cross-camera hard negative prototypes $K_2$}}
Figure \ref{fig_k1_k2_subcam} (b) further proves the effectiveness of mining multiple hard negative prototypes. Increasing the value of K2 from 1 to 80 brings steady performance boost, showing that considering more cross-camera hard prototypes is indeed beneficial for the model to better overcome camera style influence. When increasing K2 to 90, the accuracy drops a bit, indicating that top 50 to 70 hard negative prototypes are diverse enough and suffice for our loss optimization.

\begin{table}[hb]
\centering
\caption{Analysis on the proposed sub-camera environment split strategy. Only $\mathcal{L}_{Intra1}$ or $\mathcal{L}_{Intra1}^{'}$ is utilized.}
\scalebox{0.9}{
\begin{tabular}{c || cc | cc} 
\hline
\multirow{2}{*}{\shortstack{Sub-camera \\Env. Split}} &  \multicolumn{2}{c|}{Market-SCT}  & \multicolumn{2}{c}{MSMT-SCT}\\
\cline{2-5}        & R1  & mAP            & R1  & mAP  \\ 
 \hline
No             & 81.4 & \textbf{63.5}             & 45.3 & 20.6    \\
Yes             & \textbf{81.7} & 62.9             & \textbf{51.1} & \textbf{23.4}    \\
\hline
\end{tabular}
}
\label{ablation_table2}
\end{table}

\subsubsection{\textbf{Analysis on the sub-camera environment split. }}
The purpose of sub-cameras split is to create same-style environments for style-varying cameras, so that intra-environment learning can better avoid the interference of image style. In Table \ref{ablation_table2} and Figure \ref{fig_k1_k2_subcam} (c), we investigate the effectiveness of this strategy. First, Table \ref{ablation_table2} shows that intra sub-camera learning results in a similar or slightly lower accuracy on Market-SCT, compared to intra-camera learning. A possible reason is that for Market-SCT, each camera view is already style-consistent, so further splitting the camera into sub-cameras won't benefit much. On the other hand, generating sub-cameras has a positive impact on MSMT-SCT. Compared to intra-camera loss, intra sub-camera loss leads to much higher performance, improving rank-1 by $5.8\%$.

The results in Figure \ref{fig_k1_k2_subcam} (c) also back up this conclusion. When splitting into 23 sub-cameras and performing intra sub-camera learning, rank-1 and mAP are improved by $5.7\%$ and $2.2\%$ respectively. Further splitting into more sub-cameras brings consistent or even better accuracy. This is because MSMT-SCT dataset has much more diverse styles within camera, and attending to the intra-camera style variation let the model focus on learning ID-specific semantic feature instead of spurious camera feature.

\begin{figure}[h]
\centering
\begin{subfigure}{0.18\textwidth}
\centering
\includegraphics[width=1.0\textwidth]{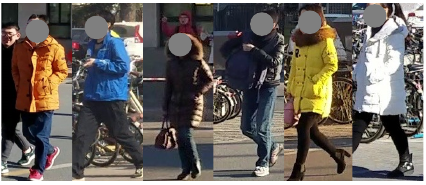} 
\caption{Sub-camera $1\_1$}
\end{subfigure}
\quad
\begin{subfigure}{0.18\textwidth}
\centering
\includegraphics[width=1.0\textwidth]{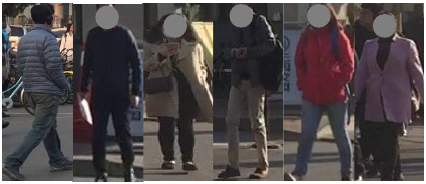} 
\caption{Sub-camera $1\_2$}
\end{subfigure}
\\
\begin{subfigure}{0.18\textwidth}
\centering
\includegraphics[width=1.0\textwidth]{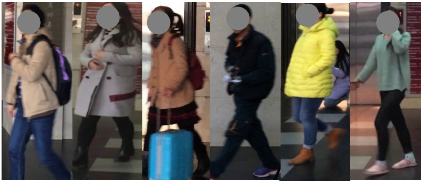} 
\caption{Sub-camera $2\_1$}
\end{subfigure}
\quad
\begin{subfigure}{0.18\textwidth}
\centering
\includegraphics[width=1.0\textwidth]{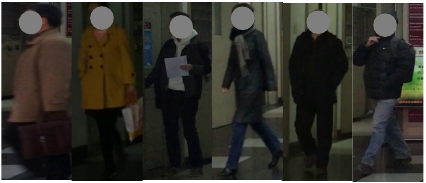} 
\caption{Sub-camera $2\_2$}
\end{subfigure}
\caption{Examples of our sub-camera split result. Images in the same row are split from the same camera view. }
\label{fig_vis_subcam}
\end{figure}

To better understand the effect of sub-camera environment split, Figure \ref{fig_vis_subcam} shows two examples of our sub-camera split result on MSMT-SCT. Each row shows the example images from two sub-cameras split from a single camera view. The first row depicts two sub-cameras split from an outdoor camera. We observe that sub-camera $1\_1$ aggregates images of a brighter image style, with bicycles in the background. Sub-camera $1\_2$ have a darker style, and image backgrounds are more empty. The second row are images from an indoor camera. Similar to the observations in the first row, images in sub-camera $2\_1$ and sub-camera $2\_2$ exhibit warmer and cooler hues respectively. And images within the same sub-camera environment generally have similar style or background. The examples show that sub-cameras are split by image style or background context, creating more consistent sub-camera environments for invariance learning.

\subsubsection{\textbf{Analysis on cross-camera identity overlap ratio.}}
ISCS re-ID strictly requires each training identity to appear only in one camera view. In practical applications, there might be chances that a portion of the identities appear in multiple cameras. To investigate how well our method can cope with such case, we perform experiments on Market-SCT with different ratios of identities appearing in multiple cameras. At a given overlapping ratio, we randomly select part of the identities from the original Market-1501 training set, and add their cross-camera images into the Market-SCT training set. Here, images of the same identity but from different cameras are assigned different ID labels. 

\begin{figure}[h]
\centering
\includegraphics[width=0.45\textwidth]{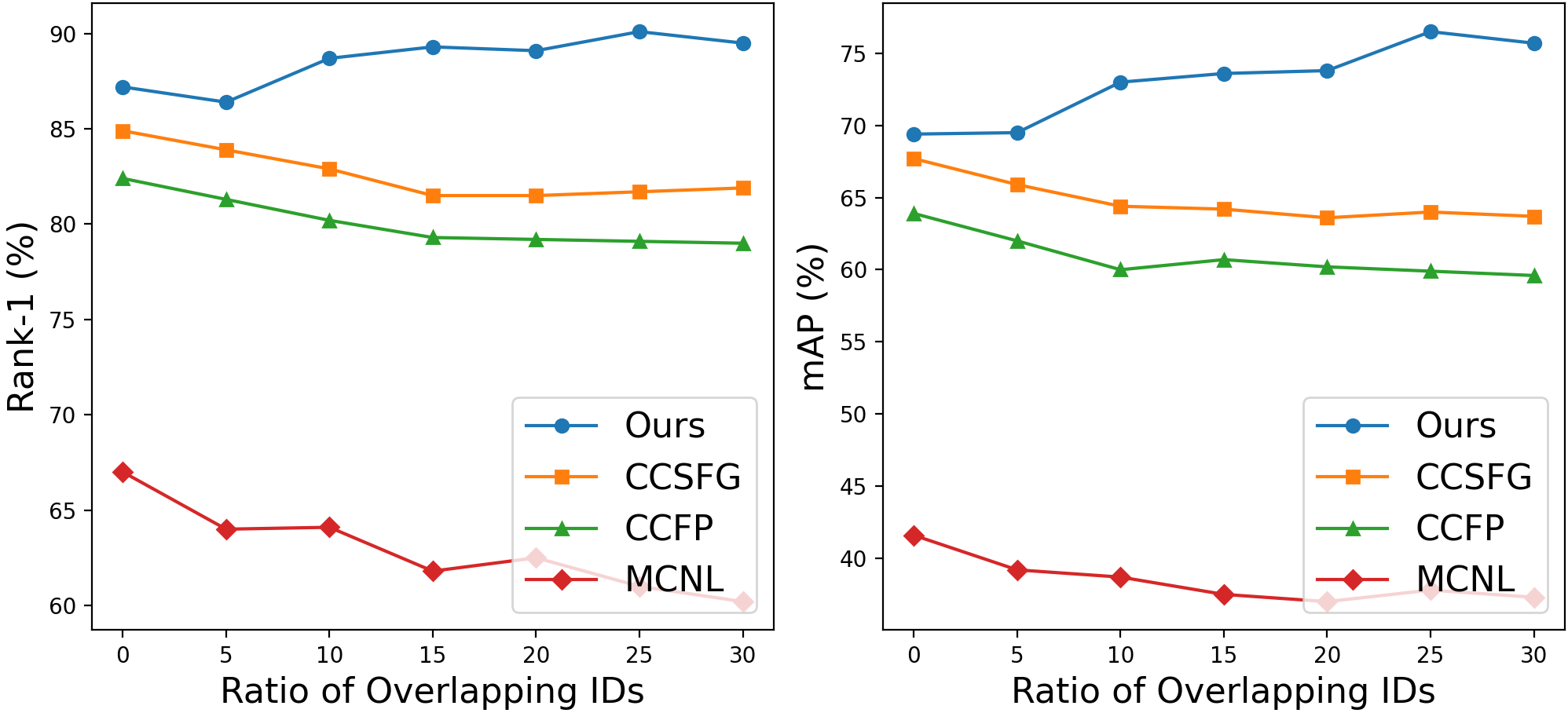} 
\caption{Model accuracy at different ratio of identity overlap. Experiments are performed on Market dataset.}
\label{fig_overlap_ratio}
\end{figure}

The results are shown in  Figure \ref{fig_overlap_ratio}. When identity overlap ratio increases from 0 to 30, MCNL, CCFP and CCSFG all experience a decrease in accuracy. However, our proposed method is able to perform stably well, and the accuracy even increases a bit when the overlap ratio is raised. Part of the reason is the intra-camera learning component in our method. The intra-camera learning performs independent contrastive learning under each camera, therefore is not negatively influenced by cross-camera re-appeared identities. Rather, the cross-camera overlapped identities may help our model at global ID discrimination due to more diverse identities being available for intra-camera contrast. Overall, the results prove the stability and robustness of our method in coping with possible identity overlap.

\begin{table*}[ht]
\centering
\caption{Comparison with state-of-the-art methods. IICI: our proposed method in this work. IICI(Res-Nonlocal): IICI with ResNet50-Nonlocal as backbone. IICI(Res-Nonlocal-DeTR): IICI with ResNet50-Nonlocal as backbone, with additional DeTR branch. IICI(Res-IBN): IICI with ResNet50-IBN as backbone. IICI(ViT-S): IICI with ViT-S as backbone.}
\scalebox{0.95}{
\begin{tabular}{c||c|cccc|cccc}
\hline  
\multirow{2}{*}{Methods} & \multirow{2}{*}{Reference} &  \multicolumn{4}{c| }{Market-SCT}  & \multicolumn{4}{c}{MSMT-SCT}  \\
\cline{3-10}&  & R1 & R5 & R10 & mAP   & R1 & R5 & R10  & mAP  \\ 
\hline
PCB~\cite{sun2018beyond}                 & ECCV18                & 43.5 & - & - & 23.5                   & - & - & - & -     \\
MGN-IBN~\cite{wang2018mgn}        & ACMMM'18         & 45.6 & 61.2 & 69.3 & 26.6         & 27.8 & 38.6 & 44.1 & 11.7    \\ 
BoT~\cite{luo2019trick}                      & CVPRW19            & 54.0 & 71.3 & 78.4 & 34.0         & 20.4 & 31.0 & 37.2 & 9.8    \\
AGW~\cite{9336268}                          & TPAMI21              & 56.0 & 72.3 & 79.1 & 36.6         & 23.0 & 33.9 & 40.0 & 11.1     \\
\hline
HHL~\cite{zhong2018generalizing}      & ECCV18             & 65.6 & 80.6 & 86.8 & 44.8         & 31.4 & 42.5 & 48.1 & 11.0     \\
MMD~\cite{long2015mmd}                & ICML15                & 67.7 & 83.1 & 88.2 & 44.0         & 42.2 & 55.8 & 61.4 & 18.2    \\
CORAL~\cite{sun2016coral}                & ECCV16              & 76.2 & 88.5 & 93.0 & 51.5         & 42.6 & 55.8 & 61.5 & 19.5    \\
\hline
MCNL~\cite{tian2020single}                & AAAI20               & 67.0 & 82.8 & 87.9 & 41.6         & 26.6 & 40.0 & 46.4 & 10.0     \\
CCFP~\cite{ge2021ccfp}                     & ACMMM'21           & 82.4 & 92.6 & 95.4 & 63.9        & 50.1 & 63.3 & 68.8 & 22.2   \\
CCSFG~\cite{Wu2022ccsfg}                & CVPR22                & 84.9 & 94.3 & 96.2 & 67.7        & 54.6 & 67.7 & 73.1 & 24.6     \\
\rowcolor{mygray}
IICI (Res-Nonlocal)                     & This work                 & 86.0 & 94.6 & 96.3 & 68.4                & 56.2 & 68.7 & 73.9 & \textbf{26.0}     \\
\rowcolor{mygray}
IICI (Res-Nonlocal-DeTR)                     & This work                     & \textbf{87.2} & \textbf{94.7} & \textbf{96.8} & \textbf{69.4}                        & \textbf{56.9} & \textbf{69.2} & \textbf{74.2} & 24.6     \\
\rowcolor{mygray}
IICI (Res-IBN)                     & This work                     & 86.3 & 94.7 & 96.5 & 69.7                         & 58.3 & 71.1 & 76.2 & 28.0     \\
\rowcolor{mygray}
IICI (ViT-S)                     & This work                     & 91.4 & 96.9 & 98.2 & 79.9                        & 65.8 & 77.2 & 81.6 & 35.5     \\
\hline
\end{tabular}
}
\label{compare_SOTA_table}
\end{table*}

\subsection{Comparison with State-of-the-arts}

We compare our proposed method with state-of-the-art methods for ISCS re-ID task, and present the results in Table \ref{compare_SOTA_table}. The compared methods are: 
1) State-of-the-art fully-supervised methods trained on Market-SCT or MSMT-SCT, including PCB~\cite{sun2018beyond}, MGN-IBN~\cite{wang2018mgn}, BoT~\cite{luo2019trick} and AGW~\cite{9336268}.
2) Domain adaptation methods HHL~\cite{zhong2018generalizing}, MMD~\cite{long2015mmd} and CORAL~\cite{sun2016coral}.
3) State-of-the-art ISCS re-ID methods MCNL~\cite{tian2020single}, CCFP~\cite{ge2021ccfp}, CCSFG~\cite{Wu2022ccsfg}, and our proposed method IICI. Specifically, we report the accuracy of our method under different backbones such as ResNet50 and ViT-S.

\subsubsection{\textbf{Comparison with SoTA methods.}} First, by looking at the results of fully-supervised methods, we can see that previous methods designed for general multi-camera re-ID suffer from accuracy drop when training set is camera-isolated. Even the best-performing AGW~\cite{9336268} lags behind methods specifically designed for ISCS re-ID by a large margin. This is because those methods typically assume each person to appear in multiple cameras in training set. Their results proves that the existence of cross-camera positive pairs is a key in high-performance fully supervised re-ID.

Next, alignment-based domain adaptation methods perform a lot better than fully-supervised re-ID methods. On Market-SCT dataset for example, CORAL~\cite{sun2016coral} even outperforms MCNL, the latter especially designed for ISCS re-ID. This is possibly because while aligning different domains, those domain adaptation methods implicitly achieves certain domain invariance. When applied on ISCS re-ID datasets, they are somewhat robust to style variation in different cameras, thus achieve a promising performance. 

Compared to fully-supervised and domain-adapted counterparts, methods designed for ISCS re-ID task generally have higher performances. CCSFG~\cite{Wu2022ccsfg} holds the best accuracy among previous ISCS methods, improving CORAL~\cite{sun2016coral} by $8.7\%$ and $12\%$ in rank-1 on Market-SCT and MSMT-SCT. Nevertheless, our proposed method IICI surpasses CCSFG and reaches a new state-of-the-art. Using the same backbone(Res-Nonlocal-DeTR) as CCSFG and CCFP, our method achieves $87.2\%$ rank-1 accuracy on Market-SCT, outperforming CCSFG by $2.3\%$. On MSMT-SCT, the improvement is also similar. The comparison proves the effectiveness of our proposed method for ISCS re-ID task. Without adding additional sub-networks, our method is able to reach higher performance, indicating the suitability of exploiting invariance learning for isolated camera supervised re-ID.

\subsubsection{\textbf{Our method with different backbones.}} Finally, we also report the performance of our method using other backbones, including ResNet-IBN and ViT-S. ResNet-IBN is commonly adopted for re-ID task. Here we experiment with ResNet-IBN as backbone, and provide the results for comparison. As shown in Table \ref{compare_SOTA_table}, IICI(Res-IBN) using a pure ResNet architecture achieves comparable performance to IICI(Res-Nonlocal-DeTR). On MSMT-SCT, IICI(Res-IBN) reaches $58.3\%$ rank-1 and $28.0\%$ mAP accuracy, improving IICI(Res-Nonlocal-DeTR) by a clear margin. In addition, when using more advanced backbone ViT-S, the accuracy is significantly boosted on both two datasets, showing that transformer architecture is indeed more powerful, and that our method is compatible and suitable under different backbones. It should be noted that when training ViT-S on MSMT-SCT, we use the sub-camera split result by ResNet-IBN, since we empirically find ResNet-IBN to have a better ability at generating sub-camera environments with its low-layer features.

\section{Conclusion}
In this paper, we have proposed a new method for isolated camera supervised re-ID. Exploiting the variance in training data, our method learns both intra-camera and inter-camera invariance under a unified framework. For intra-camera learning, we avoid the style shortcut by prototypical contrastive learning under style-consistent environment, assisted by augmentation-invariant contrast. For inter-camera learning, we confront the camera style variation and improve the existing multi-camera loss by considering topK hardest negatives and comparing with global prototypes. The resulting model is robust to style variation and gains good ID discrimination ability, achieving state-of-the-art performance on multiple benchmark datasets.

\section*{Acknowledgement}
We would like to thank Haokun Chen for the useful discussion and feedback.

\bibliographystyle{ACM-Reference-Format}
\balance
\bibliography{reference_v1}

\end{document}